\documentclass[letterpaper, 10 pt, conference]{ieeeconf}  

\IEEEoverridecommandlockouts                              

\usepackage{graphicx}
\usepackage{cite}
\usepackage{amsmath,amssymb,amsfonts}
\usepackage{algorithmic}
\usepackage{textcomp}
\usepackage{xcolor}
\usepackage{float}
\usepackage{subfigure}
\usepackage{balance}
\usepackage{stfloats}
\usepackage{wrapfig}
\usepackage{url}
\usepackage{hyperref}
\usepackage[ruled,vlined]{algorithm2e}

\title{\LARGE \bf
Manipulate as Human: Learning Task-oriented Manipulation Skills by Adversarial Motion Priors 
}

\author{Ziqi Ma$^{1}$, Changda Tian$^{2}$ and Yue Gao$^{3*}$
\thanks{This work was supported by the National Natural Science Foundation of China (Grant No. 92248303 and No. 62373242), the Shanghai Municipal Science and Technology Major Project (Grant No. 2021SHZDZX0102), and the Fundamental Research Funds for the Central Universities.}
\thanks{$^{1}$Ziqi Ma is with the ParisTech Elite Institute of Technology, Shanghai Jiao
Tong University, Shanghai, P.R. China,
        {\tt\small  ziqi\_ma0605@163.com}}%
\thanks{$^{2}$Changda Tian is with Department of Automation, Shanghai Jiao
Tong University, Shanghai, P.R. China,
        {\tt\small  dada@ics.forth.gr}}%
\thanks{$^{3}$Yue Gao is with MoE Key Lab of Artificial Intelligence and AI Institute, Shanghai Jiao Tong University, Shanghai, P.R. China,
        {\tt\small yuegao@sjtu.edu.cn}}%
\thanks{$^*$ Corresponding author}
}

\begin{document}

\maketitle
\thispagestyle{empty}
\pagestyle{empty}

\begin{abstract}
In recent years, there has been growing interest in developing robots and autonomous systems that can interact with human in a more natural and intuitive way. One of the key challenges in achieving this goal is to enable these systems to manipulate objects and tools in a manner that is similar to that of humans. In this paper, we propose a novel approach for learning human-style manipulation skills by using adversarial motion priors, which we name HMAMP. The approach leverages adversarial networks to model the complex dynamics of tool and object manipulation, as well as the aim of the manipulation task. The discriminator is trained using a combination of real-world data and simulation data executed by the agent, which is designed to train a policy that generates realistic motion trajectories that match the statistical properties of human motion.
We evaluated HMAMP on one challenging manipulation task: hammering, and the results indicate that HMAMP is capable of learning human-style manipulation skills that outperform current baseline methods. Additionally, we demonstrate that HMAMP has potential for real-world applications by performing real robot arm hammering tasks.
In general, HMAMP represents a significant step towards developing robots and autonomous systems that can interact with humans in a more natural and intuitive way, by learning to manipulate tools and objects in a manner similar to how humans do. The code is online: \url{https://github.com/ZiqiLoveSunshine/Manipulate_as_Human-AMP}
\end{abstract}

\section{Introduction}

Manipulating tools with robot arms is a long-standing area of study in the field of robot intelligence. To effectively manipulate an object with a tool and achieve a specific goal, robots must develop a comprehensive understanding of the environment based on sensor data and then perform intricate physical interactions with targets. Several prior efforts have focused on data-driven methods that aim at learning reliable tool representations for manipulation tasks. In particular, the application of end-to-end deep neural networks has gained popularity for acquiring such representations \cite{fang2020learning, ainetter2021end, kalashnikov2018qt}. These methods allow for the generation of latent tool object representations through end-to-end neural networks and extensive datasets, eliminating the need for manually crafting stage pipelines and defining object features. However, these black-box methods often lack of interpretability and compactness, and they may not fully account for subsequent manipulation processes, as they do not consider actions beyond the successful tool grasp.

\begin{figure}
    \centering
    \includegraphics[width=0.5\textwidth]{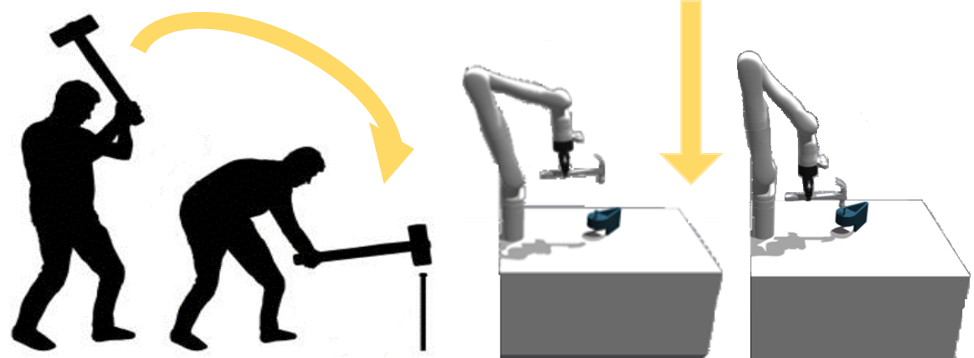}
    \caption{Difference of hammering between humans and robots. When humans hammer the nail, they swing the hammer in the opposite direction of striking in order to stock energy, while robots only focus on the achievement of task and ignore this important action}
    \label{stru_with_strength}
\end{figure}

Some works try to use keypoints to define the tool and environment in order to mathematically formulate the manipulation task. In the works of Qin \emph{et al.}\cite{qin2020keto} and Manuelli \emph{et al.} \cite{manuelli2022kpam}, they express the tool and the task in keypoints, and devise optimization algorithms to manipulation actions. By acquiring tool keypoints through supervised or reinforcement learning and defining task keypoints in the environment, these approaches formulate Quadratic Programming problems to generate robot movement trajectories within the task context. Another recent study by Turpin \emph{et al.} \cite{turpin2021gift} also adopts keypoints to represent tool objects. However, they employ reinforcement learning to predict tool affordances using a complexly designed reward function.

Nevertheless, previous studies focus on achieving task completion in complex manipulation scenarios as they often overlook whether the manipulation trajectory of robot is similar to that of human being. For instance, when faced with intricate tasks, such as hammering a nail, these methods typically provide a straightforward policy. After grasping a hammer, they position it directly on the nail and strike down until contact occurs. However, human hammering involves a distinct action: the buildup of kinetic energy by swinging the hammer in the opposite direction before striking above the nail's position. This natural human-style approach stores energy in the hammer during the swing and releases it upon impact with the nail. Such nuanced behavior is challenging to learn using conventional optimization methods because encoding the swing action with constraints in an objective equation proves difficult. The works in \cite{liu2019mirroring, zhang2022understanding, doi:10.1126/scirobotics.aay4663} use various sensors to capture the effects that human conduct on the tool and on the oriented objects to generalize the use of tool from human to robot, these methods provide a preview by learning the human movement on the physical effects, however, the inputs that they use depends on tactile data, it is expensive to gain in the normal life and will limit the propagation of the method.

Recent advances have seen a surge in deep reinforcement learning algorithms for manipulation tasks\cite{johns2021coarse,wang2023mimicplay,zorina2021learning} with easy data gain, highlighting their potential for teaching robots tool-based tasks. In imitation learning (IL), there has been progress, with supervised training using human-teleoperated demonstrations\cite{zhang2018deep} or hand-manipulated trajectories\cite{johns2021coarse}. The research in \cite{zorina2021learning} introduces a robot tool manipulation strategy using human manipulation videos. They create a simulation environment aligned with the guidance video, and calculate robot states with guided policy samples and trajectory optimization. Although effective, this approach involves solving optimization problems for each aligned environment, which consumes considerable time and computing resources.

In order to teach a robot to learn human manipulation skills in a more natural and intuitive way, we combine Adversarial Motion Priors (AMP) within a reinforcement learning problem. AMP\cite{peng2021amp} is a cutting-edge approach first appearing in computer graphics, it uses an adversarial network to learn a "style" with a reference motion dataset. The reward function involves a style reward that encourages the agent to replicate similar trajectories to those in the dataset and a task reward that assesses whether the agent achieves the task while mimicking the motion style. We also adapt style and task rewards to teach a robot arm human-like tool manipulation skills using demonstration video clips. Our approach involves competitive training between a policy network and an adversarial network. The policy network is trained using both task-specific and adversarial rewards to generate a policy that accomplishes the task with human style. The adversarial network acts as a discriminator, determining the origin of state transitions and providing a reward that effectively motivates the training of an agent. Due to the use of adversarial motion prior in guiding the robot to learn human-like manipulation skills, we name the method HMAMP. 
The contribution of our work is that: 
\begin{itemize}
    \item We introduce an implementation of task-oriented reinforcement learning combined with style in manipulation domain, and evaluate its performance on the task hammering.
    \item We provide an idea where the training data is easily acquired to learn a robot tool manipulating policy in human style.
    \item We construct an environment of tool manipulation in simulation and verify that the HMAMP is also useful in the real world.
\end{itemize}

\begin{figure*}[!t]
    \centering
    \includegraphics[width=0.96\textwidth]{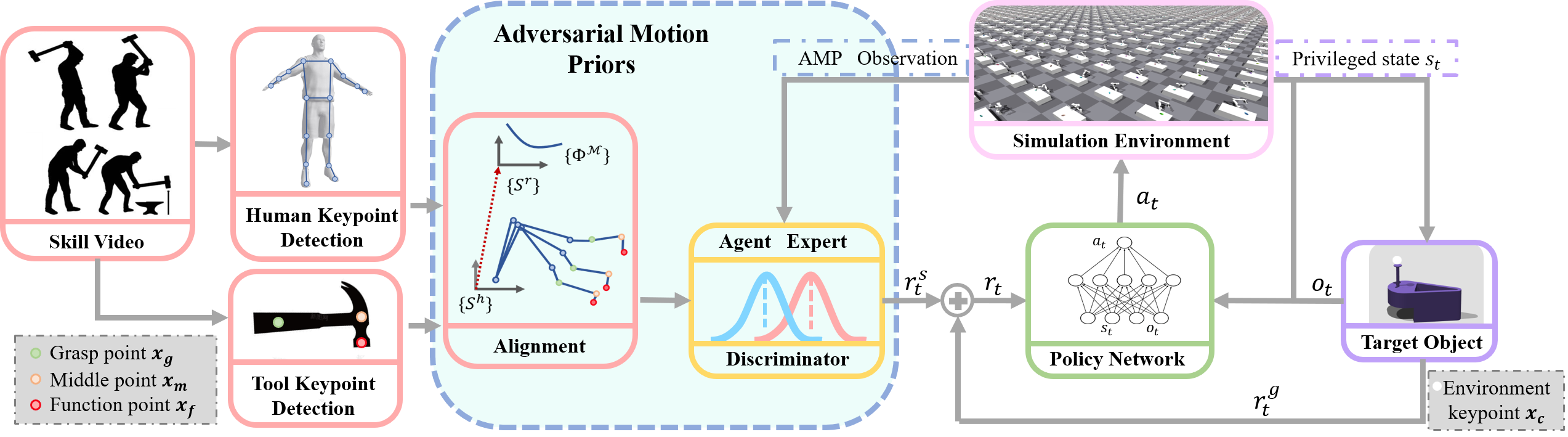}
    \caption{Framework of HMAMP. With human manipulation video clips, we extract the keypoints of human arm and manipulation tools. Then we do keypoints alignment between robot arm in simulation and real world human motion clips. The AMP Discriminator is to discriminate whether an action sequence is a real human expert motion or generated by the policy network. The AMP reward and task reward for manipulation task is added to be the total reward for RL training}
    \label{framework}
\end{figure*}


\section{Related work}
\subsection{Manipulation of Tools}

Tools use has been a fundamental issue in cognitive science studies that seeks to comprehend the nature of intelligence\cite{book,STAMANT20081199,VANLAWICKGOODALL1971195}. To enable robots to perform complex tasks that require the use of tools, many advanced studies have focused on recognizing affordance-specific features on tool objects\cite{9364360,chen2022learning}, which describes the potential physical interaction between object and manipulator and associates the regional part to planned sequential actions. These approaches have been successful in equipping robots with the capability to understand how objects may serve different purposes. In addition to recognizing the features of tool objects, learning and planning are essential components in the manipulation of tools, as demonstrated in various studies\cite{qin2020keto,turpin2021gift,DBLP:journals1}. 
Researchers have also explored methods of incorporating real-time feedback and environmental factors to improve the accuracy and precision of tool grasping\cite{RIBEIRO2021103757,ALSHANOON2022104041}. Some studies aim to identify suitable tools for a certain task, they learn an embedding knowledge by DNN model between grasping tool, desired action and target goal\cite{DBLP:journals/corr/abs-2106-02445,9629256}.

\subsection{Learning from Human Videos}
Plenty of recent research has explored utilizing human videos to improve the efficiency of RL in robots. Some works use data from the egocentric view to enable robots learning human skills\cite{xiong2022robotube,nair2022rm}.  However because of the variety of data source and the slight difference of view, it becomes difficult to use the pre-trained representation in a specific manipulation task. 
In order to mend the domain gap, another types of work utilizes the in-domain human demonstration, where the sequence of human pose is recorded by motion capture\cite{DBLP:journals/corr/abs-2008-11200} or by a view from the third person\cite{DBLP:journals/corr/abs-2101-07241}. Data of this kind has a narrower disparity between the human and robot domains, which makes it possible to construct efficient reward function for training imitation learning algorithms.  Instead of extracting and re-targeting the whole human pose from the video, we focus on the motion of parts of important joints and the motion of the tool, which allows a flexible transfer from human morphology to robot morphology. By cooperating the task-oriented approach with an adversarial motion prior, we enables our system to learn the movement of robot arm using unstructured motion data.


\subsection{Generative Adversarial Imitation Learning}

Generative adversarial imitation learning (GAIL)\cite{ho2016generative} is inspired by the idea developed for generative adversarial networks (GAN)\cite{goodfellow2020generative}. It aims to train generators that learn policies matching the trajectory distribution of the dataset, meanwhile, the discriminators serve as the reward functions to judge whether the generated behaviors look like the demonstrations. By using a small number of demonstration data from experts, GAIL learns both the unknown environment's policy and reward function. Although these methods have demonstrated success in low-dimensional domains\cite{ho2016generative}, their performance in high-dimensional tasks is not outstanding. 
Recently, Peng \emph{et al.} \cite{peng2021amp} have introduced Adversarial Motion Priors (AMP), which integrates tasks goals with generative adversarial imitation learning. This allows simulated agents to perform high-level tasks by learning to imitate behaviors from extensive motion dataset. Escontrela \emph{et al.} \cite{escontrela2022adversarial} also apply this adversarial technique with a limited number of reference motion clips to learn locomotion skills for legged robots. In manipulation areas, we apply this technique to guide robots to interpret and perform behavior shown by demonstration and we show that the agents have more flexibility to perform more natural and feasible behaviors. 


\section{Learning Tool Manipulation Policy}
We model the problem of learning human-like tool manipulating skills as a Markov Decision Process (MDP).The goal of the reinforcement learning is to find the parameter $\theta$ that optimize policy $\pi_{\theta}$ in which the expected discounted return is maximal
$J(\theta)=\mathbb{E}_{\pi_{\theta}}\left[\sum_{t=0}^{T-1} \gamma^{t} r_{t}\right]$.
We then propose the mathematical abstraction of tool manipulation task and introduce the design of reward function.

\subsection{Tool Keypoint Definition}
For each manipulation task with tool, we make the assumption that the tool objects interact with the environment containing one or more target objects. Inspired by\cite{qin2020keto}, in the framework of HMAMP, we define some keypoints in each task manually, which consists of a set of keypoints on the tool $\boldsymbol{K_o}$ and a set of keypoints in the environment $\boldsymbol{K_e}$. Specifically, we consider $\boldsymbol{K_o} =\left[ \boldsymbol{x_g}, \boldsymbol{x_f}, \boldsymbol{x_m} \right]$, where $\boldsymbol{x_g}$ characterises the grasping position on the tools and function point,  $\boldsymbol{x_f}$ characterises the functional part that make contact with the target object and $\boldsymbol{x_m}$ represents an auxiliary point that aides to determine the direction of tool. Environment keypoints is represented as $\boldsymbol{K_e} =\left[ \boldsymbol{x_c}\right]$, where $\boldsymbol{x_c}$ denotes the position where the target interacts with the tool. The keypoints of tools and environment are mostly used to determine the goal reward which needs to be optimized in a reinforcement learning problem.

\subsection{Rewards Design}
As mentioned in \cite{peng2021amp}, the total reward $r_t$ is determined as a goal reward $r^g_t$, which is specific to the task, helps to describe the completion of task, and a style reward $r^s_t$ that evaluates whether the behaviors produced by the agent is similar to the behaviors produced via the distribution of reference motion set. The more similar the behaviors are, the higher value the style reward is. The proportion between style reward and goal reward is adjusted manually before training.
\begin{equation}
r_t = \alpha^g r^g_t + \beta^s r^s_t
\label{e1}
\end{equation}

\subsubsection{Goal Reward about Task}
The goal reward related to the task and must be designed specifically.
For the tool manipulation tasks where there is a contact with environment, we design the goal reward $r^g_t$ composed by two terms at one instance $t$.

\begin{equation}
\begin{aligned}
r^g_t &= \omega^f r^f_t + \omega^d r^d_t \\
r^f_t &= \left\{
    \begin{array}{rcl}
    \lVert F^{s}_t(\boldsymbol{x}_f,\boldsymbol{x}_c)\rVert /F^d
     & & {F^{s}_t \leq F^d} \\
    1 \quad \quad \quad \quad & & {F^{s}_t > F^d}
    \end{array}\right. \\
r^d_t &= 1- \tanh{(\left\lVert \boldsymbol{x}_f - \boldsymbol{x}_c\rVert _t\right)}
\end{aligned}
\end{equation}

The first term $r_t^f$ stimulates task completion, which is defined with the contact force detected on the target object that is exerted by the tool, the force is captured by force sensor in the simulation. The detected force is encouraged to converge towards the force $F_d$ that we desire. This reward has values only when contact occurs and after the detection of contact force, the policy terminates. The second term $r_t^d$ is defined between tool function point $\boldsymbol{x}_f$ and environment target point $\boldsymbol{x}_c$ to guide to exploit a policy that minimizes the distance between tool and target. The utilise of $\tanh$ function aims to bound the reward to $[0, 1]$. These terms are weighted with manually-specified coefficients $\omega^f$ and $\omega^d$, however, the magnitude of second term is much lower than the first.

\subsubsection{Style Reward with Motion Prior}
As the idea mentioned in \cite{peng2021amp}, we define a discriminator $D_{\phi}$ as a neural network with parameter $\phi$, the discriminator is trained to distinguish whether a transition $(s, s')$ is a fake one produced by an agent or a true one sampled from a real motion distribution $d^\mathcal{M}$.

We update the objective of discriminator as: 

\begin{equation}
\begin{aligned}
    \underset{\phi}{\operatorname{arg min}}\quad &\mathbb{E}_{d^\mathcal{M}(s,s')}\left[\left(D_{\phi}(s,s')-1\right)^2\right] \\+&\mathbb{E}_{d^\pi(s,s')}\left[\left(D_{\phi}\left(s,s'\right)+1\right)^2\right] \\
    +&\dfrac{w^{gp}}{2}
    \mathbb{E}_{d^\mathcal{M}(s,s')} \left[\lVert\nabla_{\phi}D_{\phi}\left(s,s'\right)\rVert^2\right]
\end{aligned}
\label{D-objective}
\end{equation}

The former two terms serve to motivate the discriminator to differentiate between the input state derived from a policy and the input state derived from the reference motion data. They are proposed in LSGAN\cite{DBLP:journals/corr/MaoLXLW16} to solve the challenge of vanishing gradients caused by the standard GAN objective function where a sigmoid cross-entropy loss function is usually used. LSGAN prefer to optimize $\chi^2$ divergence between the reference distribution and the policy distribution, which may alleviate the mode collapse problem and lead to stable performance during the training process\cite{article}. The last term in \eqref{D-objective} is a gradient penalty term to penalize nonzero gradients on samples, which may avoid oscillations and improve training stability. The $w^{gp}$ in the formula is a coefficient adjusted manually. 

The style reward is then defined by:
\begin{equation}
    r^s_t = \max\left[0, 1-\gamma^d (D_{\phi}(s,s')-1)^2 \right]
    \label{e4}
\end{equation}
with the additional offset and scale, the style reward is bounded between $[0, 1]$.

The training process of policy and discriminator is shown in Fig. \ref{framework}. The agent steps to interact with environment and produces a state transition $(s,s')$, the observation in the environment is used to calculate the goal reward $r^g_t$. The discriminator takes the state transition from a simulated environment and from a reference motion clips to calculate the style reward $r^s_t$. In the end, the combined reward is used to optimize competitively the policy and the discriminator. The training details is demonstrated in Alg. \ref{alg:amp_learning}.

\begin{algorithm}
\caption{HMAMP: Learning Human-like Manipulation Skills by Adversarial Motion Prior}
\label{alg:amp_learning}
\begin{algorithmic}[1]
\STATE \textbf{Input}: Task-specific environment $\mathcal{E}$, Dataset of human reference motions $\mathcal{M}$
\STATE Initialize AMP discriminator $D$
\STATE Initialize policy $\pi$
\STATE Initialize reply buffer $\mathcal{B}$

\FOR{each training episode}
    \STATE Reset environment $\mathcal{E}$ to initial state
    \FOR{trajectory $i = 1,\dots,m$}
        \STATE collect trajectory $\tau_i$ with $\pi$: $\tau_i\leftarrow \{(s_t,a_t,r^g_t)^{T-1}_{t=0},s^g_T\}$
        \FOR{time step $t = 0,\dots,T-1$}
            \STATE $d_t \leftarrow D\phi(s_t,s'_t)$
            \STATE calculate $r^s_t$ by Equation \eqref{e4}.
            \STATE calculate $r_t$ by Equation \eqref{e1}.
            \STATE record $r_t$ in $\tau_i$.
        \ENDFOR
        \STATE store $\tau_i$ in $\mathcal{B}$.
    \ENDFOR
    \FOR{update step $= 1,\dots,n$}
        \STATE $b^\mathcal{M} \leftarrow$ sample batch of $K$ transitions from $\mathcal{M}$.
        \STATE $b^\pi \leftarrow$ sample batch of $K$ transitions from $\mathcal{B}$.
        \STATE update $D$ according to Equation \eqref{D-objective} using $b^\mathcal{M}$ and $b^\pi$.
    \ENDFOR
    \STATE update $\pi$ using data from trajectories $\{\tau_i\}^m_{i=1}$
\ENDFOR
\end{algorithmic}
\end{algorithm}

\section{Training}

\subsection{Data Preprocess}
The raw data used for HMAMP consists of video clips capturing manipulation skills from a third-person perspective. This type of data offers several advantages: it minimizes the domain gap between simulation and reality, and it is relatively easy to acquire, making it a practical choice for training purposes. In this work, we create the dataset which records human skill: hammering. The duration of the human motion in each video clips is less than one second, and we collect five pieces of video clips that perform the hammering movement by two persons. Then, we use the most popular keypoint detection algorithm BlazePose\cite{bazarevsky2020blazepose} to detect human joint including Hip, Elbow, Wrist, Hand and we use a CV algorithm to extract time-series of tool keypoints by manually signing them on the tool. 

To retarget human motion to robot motion, it is necessary to construct an effective transfer function that maps the human world space to the robot world space. Numerous studies\cite{6512060, GENG2011272, 7139991} have explored motion retargeting between these two domains, and any established retargeting method can be applied in this process. In our work, we adopt a simple and straightforward approach: direct mapping. While there are significant differences in topology between the human arm and the robot arm, they share corresponding joints, such as the human elbow, wrist, and hand, which align with certain robot joints and the end-effector. By mapping key human joints to their robotic counterparts, we achieve a rough but effective transfer of motion from the human domain to the robot domain.

\subsection{Model Details}
The policy that we used in HMAMP is PPO, the hidden parameters of policy network is of size [512, 256, 128] with exponential linear unit activation layers. The policy outputs the distribution from which the target joint positions are sampled with the representation of mean and standard deviation. Then, the target joint positions are fed to our customized PD controllers to compute the motor torques. The policy is trained on an observation $o_t$ derived from the state, which contains environment information such as hammer position and nail position and robot information such as joint angles, joint velocities, end-effector orientation, and previous actions. The discriminator is an MLP with hidden layers of size [1024, 512] and exponential linear unit activation layers. It takes the the orientation of robot joints and the orientation of tools as input. The value of all manually determined parameters are: $\alpha^g = 0.6, \beta^s = 0.4, \gamma^d = 0.25, \omega^f = 10^5, \omega^d = 1, \omega^{gp} = 1, F^d = 100$.

\subsection{Simulation}
\begin{figure}
    \centering
    \includegraphics[width=0.5\textwidth]{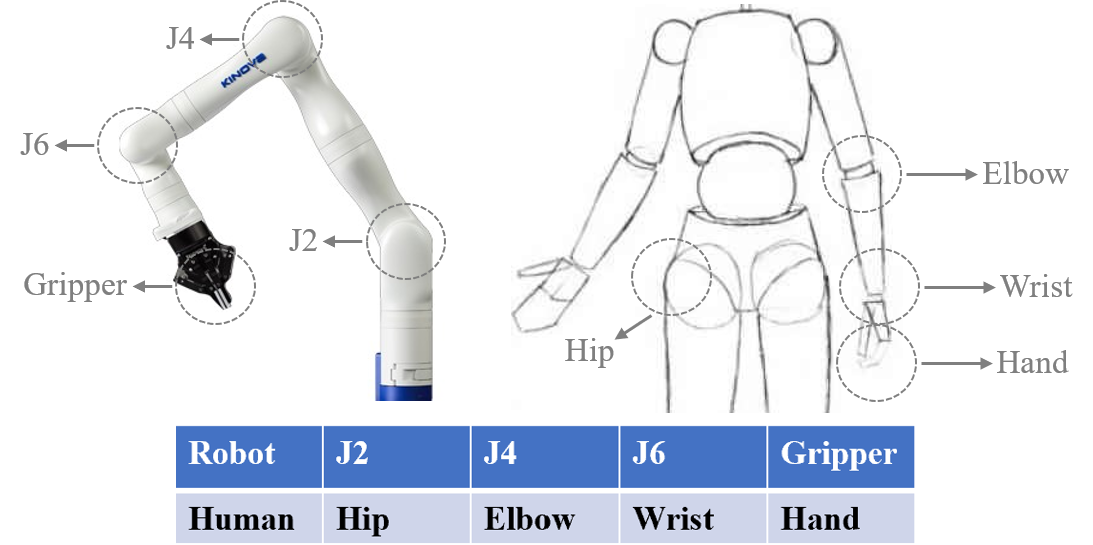}
    \caption{Direct mapping between human and robot arm. Some joints and the gripper of a Kinova Gen3 are mapped to act as human hip, elbow, wrist, and hand.}
    \label{robot-human}
\end{figure}

The robot that we choose to train the policy is Kinova Gen3 with the gripper 2f85, the correspondence joint mapping result between Gen3 and human is shown in Fig \ref{robot-human}.

We selected Isaac Gym \cite{isaac} as our simulation platform due to its ability to accelerate the RL training process using GPU resources. The policy was trained in parallel across 2,048 agents, utilizing a single NVIDIA RTX 3090 GPU. The entire training process required 11 hours of wall-clock time and spanned 60,000 training epochs. Each RL episode lasted a maximum of 152 steps, corresponding to 3 seconds of simulated time, and terminated early if the termination criteria were met. The policy operated at a control frequency of 50 Hz during the simulation.

\subsection{Termination}
An episode terminates and the next one starts when the robot satisfy the termination criteria. It contains task finish signal when the hammer knocks the nail and the collision signal when a force is detected between robot components or between the robot and the table or the hammer and the table. 

\begin{figure*}[!t]
\centering
\includegraphics[width=1\textwidth]{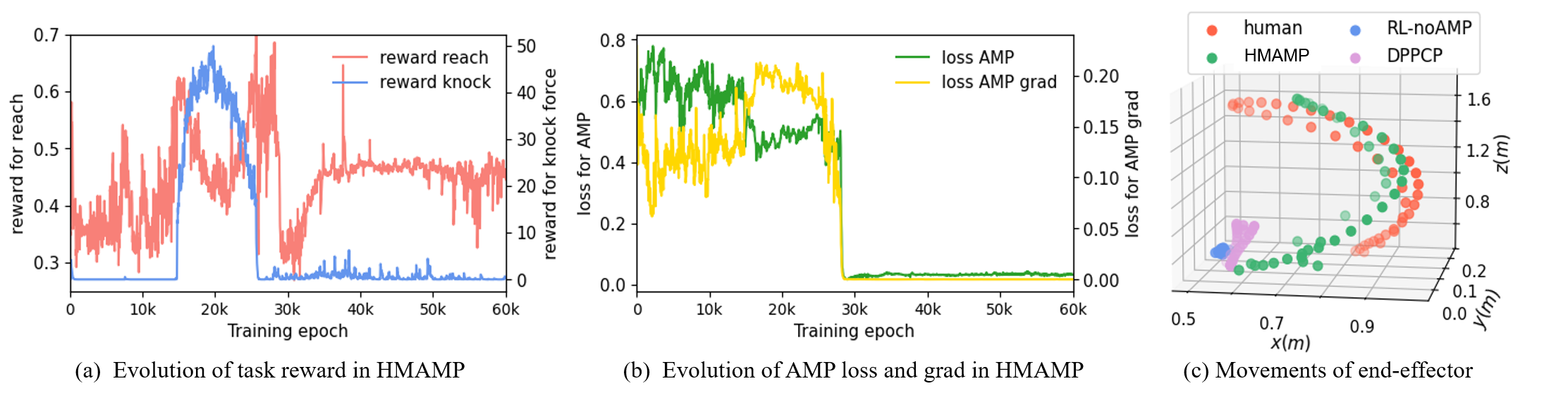}
\caption{The training process of HMAMP and the end-effector tracking comparison of HMAMP and baselines. Figure (a) shows the evolution of reach reward and knock force reward in the training process. Figure (b) shows the discriminator loss and gradient in the training process. The two figures show the confrontation and balance between style reward and goal reward. In the early stage goal reward has a strong guiding effect while in the late stage amp discriminator converges quickly, giving the trajectory of robot a human style. Figure (c) shows movement trajectory of the end-effector of the robot arm in Cartesian space. The motion trajectory obtained by HMAMP is the most similar to human expert's trajectory.}
\label{train_proc_trajectory}
\end{figure*}

\subsection{Domain Randomization and Training Process}

In order to improve the robustness of HMAMP policy and facilitate the transfer of learned policy from simulation to the real world, we apply domain randomization in the training process. In details, we randomize the coefficient of friction applied to hammer and nail to $[0.5, 1.25]$ and the joint-level PD gains to $[0.9, 1.1]$. In addition, the same observation noise as in \cite{zorina2021learning} is added during the training phase, where the Cartesian position observation noise is $\pm 0.01 m$ and joint position observation noise is $\pm 0.02 rad$.



Form Fig. \ref{train_proc_trajectory}(a)(b), the training curve shows the confrontation and balance between style reward and goal reward. The early stage goal reward has a strong guiding effect, while in the late stage, amp discriminator quickly converges, giving the movements of robot a human style.



\section{Experiment}

In this section, we present the experimental setup and results to demonstrate the effectiveness of HMAMP in learning task-oriented, human-like manipulation skills. To evaluate the performance of our method, we conducted a series of comparative experiments against two baseline approaches: a direct path-planning control policy and a reinforcement learning (RL) approach without AMP. The experimental results were recorded and analyzed quantitatively to highlight the contributions of HMAMP. The evaluation focused on three key aspects: the quality of the learned manipulation skills, task completion efficiency, and the similarity of the robot's movements to human behavior.

\subsection{Comparative Experiment in Simulation}
\subsubsection{Task Definition}

The chosen manipulation task for the experiments involves knocking a nail with a hammer. The experiments are conducted on the Isaac Gym simulation platform, using the model of a 7-dof Kinova Gen3 Arm with 2f-85 gripper to complete the task. At the start of the task, the hammer is securely grasped by the robotic arm, which begins in its initial home position, with the gripper oriented perpendicular to the platform (see Fig. \ref{real_exp}). The nail is placed arbitrarily on the manipulation platform. The objective of the task is to successfully hammer the nail while replicating a human style of movement.

\subsubsection{Baseline Methods}
\label{baseline_methods}
We compare HMAMP against the following baseline methods:
\begin{itemize}
    \item Direct Path-Planning Control Policy (DPPCP): This baseline approach determines manipulation actions using predefined path-planning strategies. In this method, proportional-derivative (PD) control is employed to generate a planned trajectory guiding the hammer from its initial position to the nail.
    \item Reinforcement Learning without AMP (RL-noAMP): This baseline approach uses a standard reinforcement learning method for the agent to acquire manipulation skills. The configuration of this approach is identical to that of HMAMP, except that the adversarial motion priors (AMP) component is removed. The training process follows the same procedure as in HMAMP.
\end{itemize}

\subsubsection{Evaluation Metrics}
To quantitatively evaluate the performance of each approach, we define the following criterias:
\begin{itemize}
    \item \textbf{Knock Impulse}: A measure of the knock effect received by the nail. It is calculated by the formula : $I = \int F_{nail}(t) \, dt$. Large Impulse means the nail receives large force at one instance.
    \item \textbf{Energy Efficiency}:  The energy used by the arm is $E = \sum_{i=1}^{n} \int \tau_i(t) \cdot \omega_i(t) \, dt$, where $n$ is the joint number of the arm. The energy efficiency can be represented as the ratio of knock impulse received by the nail and the energy cost by the arm: $\eta = I/E$
    \item \textbf{Vertical Force Ratio}: The vertical force ratio reflects how efficiently the force is applied in the vertical direction, which is critical for tasks involving hammering. Higher ratios indicate more effective and efficient nail hammering: $\text{Vertical Force Ratio} = \frac{F_{\text{vertical, nail}}}{F_{\text{total, nail}}}$
    \item \textbf{Frechet Distance}: The quantitative analysis in motion similarity between human and robot arm manipulation \cite{aronov2006frechet}. The smaller the Frechet distance between two trajectories is, the more similar their shapes are.
\end{itemize}

\subsubsection{Results in Simulation}

We implemented the two baseline control strategies, DPPCP and RL-noAMP, as described in Sec. \ref{baseline_methods}, within the simulation environment. Comparative experiments were conducted to evaluate their performance against our proposed method. Each method was tested across 10 trials, and the average values for each evaluation criterion were calculated to ensure stable and reliable performance measurements.

Table \ref{comp_res} presents a comprehensive comparison of the various approaches, highlighting key performance metrics. The HMAMP consistently outperforms both DPPCP and RL-noAMP in all evaluated aspects. Notably, it has an overwhelming advantage in the impulse received by the nail, which is the most critical task of hammering nails. From the table we can also obtains that the HMAMP is more efficient in hammering, since both the energy efficiency and vertical force ratio excel the other methods. In addition, HMAMP has the closest Frechet Distance to human manipulation trajectories, which reflects the effectiveness of our method in learning human motion styles. We can experience this more intuitively from Fig. \ref{train_proc_trajectory}(c).

\begin{table}[h!]
   \caption{Comparison results of HMAMP with baselines\label{comp_res}}
    \begin{tabular}{@{\extracolsep{\fill}}lcrrrrr}
    \hline
     Method & HMAMP& DPPCP & RL-noAMP \\
    \hline
Knock Impulse      &   4238 kg•m/s  &   2351 kg•m/s     &  2507 kg•m/s   \\
Energy Efficiency  &     1.56 s/m    &        1.34 s/m     &      1.26 s/m           \\
Vertical Force Ratio    &     0.99       &        0.99      &       0.91          \\
Frechet Distance   &     0.29     &        0.88       &       1.36         \\
    \hline
    \end{tabular}
\end{table}


The experimental results clearly demonstrate the effectiveness of the proposed HMAMP framework in learning task-oriented, human-like manipulation skills through the use of Adversarial Motion Priors (AMP). The integration of AMP significantly improves task completion efficiency, knock impulse, and energy efficiency, resulting in superior performance compared to both the direct path-planning approach and traditional reinforcement learning (RL) methods.

\subsection{Real Robot Arm Experiment}

\begin{figure*}
    \centering
    \includegraphics[width=1\textwidth]{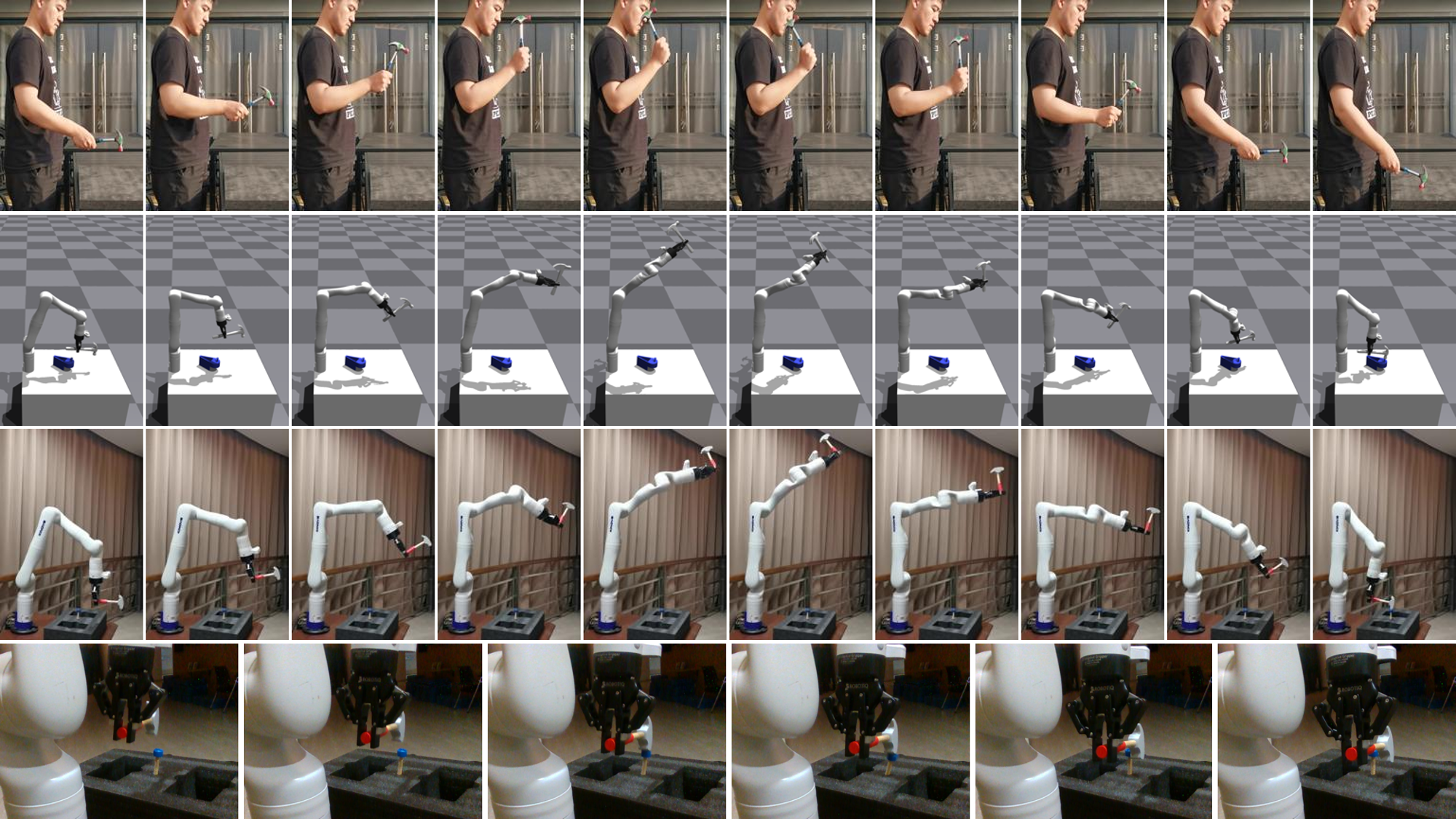}
    \caption{Experiment in simulation and real world. The first row shows human knocking motion clips that we used as motion priors. The second row shows the policy HMAMP in simulation, the hammer can successfully complete the task with the manipulation trajectory that we desired. The third row shows the HMAMP implemented in real world on Kinova Gen3, and the fourth row is the details about hammering a nail in real world.}
    \label{real_exp}
\end{figure*}

We employed the same arm and the same task as the simulation for real-world experiments. The environment setup closely resembled real-world scenarios to ensure the applicability of our approach in practical scenarios. Our proposed approach, which incorporates Adversarial Motion Priors (AMP) into Reinforcement Learning (RL), was integrated into the control system of the robotic arm. The parameters and settings were optimized based on the training results obtained in the simulation environment.
The robotic arm was tasked with performing the manipulation task using the learned skills.

Fig. \ref{real_exp} showcases the manipulation effect of HMAMP on the real robot arm. The sequence of images illustrates the robotic arm successfully completing the manipulation task with precision and human-like motion. The trajectory followed by the arm demonstrates smoothness, accuracy, energy efficiency, and human manipulation skills, confirming the benefits of incorporating Adversarial Motion Priors.

\section{Conclusion}

In this paper, we presented a novel approach named HMAMP to enable robotic arms to perform tool manipulation with human-like skills. Our method integrates Adversarial Motion Priors (AMP) with deep reinforcement learning to capture complex manipulation dynamics. By leveraging both real-world motion data and synthetic motion data generated through simulation, we demonstrated the ability of our approach to surpass existing techniques in learning human-style manipulation behaviors. The evaluation on the challenging hammering task highlighted the effectiveness of our method and its potential for real-world applications. This research bridges the gap between robotic and human capabilities, paving the way for more intuitive and natural human-robot interactions. The proposed framework serves as a foundation for future research aimed at developing robots with advanced manipulation skills, envisioning a future where machines seamlessly mimic human manipulation.


\bibliographystyle{unsrt}
\bibliography{ref.bib}

\begin{thebibliography}{10}

\bibitem{fang2020learning}
Kuan Fang, Yuke Zhu, Animesh Garg, Andrey Kurenkov, Viraj Mehta, Li~Fei-Fei, and Silvio Savarese.
\newblock Learning task-oriented grasping for tool manipulation from simulated self-supervision.
\newblock {\em The International Journal of Robotics Research}, 39(2-3):202--216, 2020.

\bibitem{ainetter2021end}
Stefan Ainetter and Friedrich Fraundorfer.
\newblock End-to-end trainable deep neural network for robotic grasp detection and semantic segmentation from rgb.
\newblock In {\em 2021 IEEE International Conference on Robotics and Automation (ICRA)}, pages 13452--13458. IEEE, 2021.

\bibitem{kalashnikov2018qt}
Dmitry Kalashnikov, Alex Irpan, Peter Pastor, Julian Ibarz, Alexander Herzog, Eric Jang, Deirdre Quillen, Ethan Holly, Mrinal Kalakrishnan, Vincent Vanhoucke, et~al.
\newblock Qt-opt: Scalable deep reinforcement learning for vision-based robotic manipulation.
\newblock {\em arXiv preprint arXiv:1806.10293}, 2018.

\bibitem{qin2020keto}
Zengyi Qin, Kuan Fang, Yuke Zhu, Li~Fei-Fei, and Silvio Savarese.
\newblock Keto: Learning keypoint representations for tool manipulation.
\newblock In {\em 2020 IEEE International Conference on Robotics and Automation (ICRA)}, pages 7278--7285. IEEE, 2020.

\bibitem{manuelli2022kpam}
Lucas Manuelli, Wei Gao, Peter Florence, and Russ Tedrake.
\newblock kpam: Keypoint affordances for category-level robotic manipulation.
\newblock In {\em Robotics Research: The 19th International Symposium ISRR}, pages 132--157. Springer, 2022.

\bibitem{turpin2021gift}
Dylan Turpin, Liquan Wang, Stavros Tsogkas, Sven Dickinson, and Animesh Garg.
\newblock Gift: Generalizable interaction-aware functional tool affordances without labels.
\newblock {\em arXiv preprint arXiv:2106.14973}, 2021.

\bibitem{liu2019mirroring}
Hangxin Liu, Chi Zhang, Yixin Zhu, Chenfanfu Jiang, and Song-Chun Zhu.
\newblock Mirroring without overimitation: Learning functionally equivalent manipulation actions.
\newblock In {\em Proceedings of the AAAI Conference on Artificial Intelligence}, volume~33, pages 8025--8033, 2019.

\bibitem{zhang2022understanding}
Zeyu Zhang, Ziyuan Jiao, Weiqi Wang, Yixin Zhu, Song-Chun Zhu, and Hangxin Liu.
\newblock Understanding physical effects for effective tool-use.
\newblock {\em IEEE Robotics and Automation Letters}, 7(4):9469--9476, 2022.

\bibitem{doi:10.1126/scirobotics.aay4663}
Mark Edmonds, Feng Gao, Hangxin Liu, Xu~Xie, Siyuan Qi, Brandon Rothrock, Yixin Zhu, Ying~Nian Wu, Hongjing Lu, and Song-Chun Zhu.
\newblock A tale of two explanations: Enhancing human trust by explaining robot behavior.
\newblock {\em Science Robotics}, 4(37):eaay4663, 2019.

\bibitem{johns2021coarse}
Edward Johns.
\newblock Coarse-to-fine imitation learning: Robot manipulation from a single demonstration.
\newblock In {\em 2021 IEEE international conference on robotics and automation (ICRA)}, pages 4613--4619. IEEE, 2021.

\bibitem{wang2023mimicplay}
Chen Wang, Linxi Fan, Jiankai Sun, Ruohan Zhang, Li~Fei-Fei, Danfei Xu, Yuke Zhu, and Anima Anandkumar.
\newblock Mimicplay: Long-horizon imitation learning by watching human play, 2023.

\bibitem{zorina2021learning}
Kateryna Zorina, Justin Carpentier, Josef Sivic, and Vladimír Petrík.
\newblock Learning to manipulate tools by aligning simulation to video demonstration, 2021.

\bibitem{zhang2018deep}
Tianhao Zhang, Zoe McCarthy, Owen Jow, Dennis Lee, Xi~Chen, Ken Goldberg, and Pieter Abbeel.
\newblock Deep imitation learning for complex manipulation tasks from virtual reality teleoperation.
\newblock In {\em 2018 IEEE International Conference on Robotics and Automation (ICRA)}, pages 5628--5635. IEEE, 2018.

\bibitem{peng2021amp}
Xue~Bin Peng, Ze~Ma, Pieter Abbeel, Sergey Levine, and Angjoo Kanazawa.
\newblock Amp: Adversarial motion priors for stylized physics-based character control.
\newblock {\em ACM Transactions on Graphics (TOG)}, 40(4):1--20, 2021.

\bibitem{book}
Crickette Sanz, Josep Call, and Christophe Boesch.
\newblock {\em Tool Use in Animals: Cognition and Ecology}.
\newblock Cambridge University Press, 03 2013.

\bibitem{STAMANT20081199}
Robert {St Amant} and Thomas~E. Horton.
\newblock Revisiting the definition of animal tool use.
\newblock {\em Animal Behaviour}, 75(4):1199--1208, 2008.

\bibitem{VANLAWICKGOODALL1971195}
Jane {Van Lawick-Goodall}.
\newblock Tool-using in primates and other vertebrates.
\newblock volume~3 of {\em Advances in the Study of Behavior}, pages 195--249. Academic Press, 1971.

\bibitem{9364360}
Ruinian Xu, Fu-Jen Chu, Chao Tang, Weiyu Liu, and Patricio~A. Vela.
\newblock An affordance keypoint detection network for robot manipulation.
\newblock {\em IEEE Robotics and Automation Letters}, 6(2):2870--2877, 2021.

\bibitem{chen2022learning}
Wenkai Chen, Hongzhuo Liang, Zhaopeng Chen, Fuchun Sun, and Jianwei Zhang.
\newblock Learning 6-dof task-oriented grasp detection via implicit estimation and visual affordance, 2022.

\bibitem{DBLP:journals1}
Adithyavairavan Murali, Weiyu Liu, Kenneth Marino, Sonia Chernova, and Abhinav Gupta.
\newblock Same object, different grasps: Data and semantic knowledge for task-oriented grasping.
\newblock {\em CoRR}, abs/2011.06431, 2020.

\bibitem{RIBEIRO2021103757}
Eduardo~Godinho Ribeiro, Raul {de Queiroz Mendes}, and Valdir Grassi.
\newblock Real-time deep learning approach to visual servo control and grasp detection for autonomous robotic manipulation.
\newblock {\em Robotics and Autonomous Systems}, 139:103757, 2021.

\bibitem{ALSHANOON2022104041}
Abdulrahman Al-Shanoon and Haoxiang Lang.
\newblock Robotic manipulation based on 3-d visual servoing and deep neural networks.
\newblock {\em Robot. Auton. Syst.}, 152(C), jun 2022.

\bibitem{DBLP:journals/corr/abs-2106-02445}
Namiko Saito, Tetsuya Ogata, Satoshi Funabashi, Hiroki Mori, and Shigeki Sugano.
\newblock How to select and use tools? : Active perception of target objects using multimodal deep learning.
\newblock {\em CoRR}, abs/2106.02445, 2021.

\bibitem{9629256}
Ming Sun and Yue Gao.
\newblock Gater: Learning grasp-action-target embeddings and relations for task-specific grasping.
\newblock {\em IEEE Robotics and Automation Letters}, 7(1):618--625, 2022.

\bibitem{xiong2022robotube}
Haoyu Xiong, Haoyuan Fu, Jieyi Zhang, Chen Bao, Qiang Zhang, Yongxi Huang, Wenqiang Xu, Animesh Garg, and Cewu Lu.
\newblock Robotube: Learning household manipulation from human videos with simulated twin environments.
\newblock In {\em 6th Annual Conference on Robot Learning}, 2022.

\bibitem{nair2022rm}
Suraj Nair, Aravind Rajeswaran, Vikash Kumar, Chelsea Finn, and Abhinav Gupta.
\newblock R3m: A universal visual representation for robot manipulation.
\newblock In {\em 6th Annual Conference on Robot Learning}, 2022.

\bibitem{DBLP:journals/corr/abs-2008-11200}
Omid Taheri, Nima Ghorbani, Michael~J. Black, and Dimitrios Tzionas.
\newblock {GRAB:} {A} dataset of whole-body human grasping of objects.
\newblock {\em CoRR}, abs/2008.11200, 2020.

\bibitem{DBLP:journals/corr/abs-2101-07241}
Haoyu Xiong, Quanzhou Li, Yun{-}Chun Chen, Homanga Bharadhwaj, Samarth Sinha, and Animesh Garg.
\newblock Learning by watching: Physical imitation of manipulation skills from human videos.
\newblock {\em CoRR}, abs/2101.07241, 2021.

\bibitem{ho2016generative}
Jonathan Ho and Stefano Ermon.
\newblock Generative adversarial imitation learning.
\newblock {\em Advances in neural information processing systems}, 29, 2016.

\bibitem{goodfellow2020generative}
Ian Goodfellow, Jean Pouget-Abadie, Mehdi Mirza, Bing Xu, David Warde-Farley, Sherjil Ozair, Aaron Courville, and Yoshua Bengio.
\newblock Generative adversarial networks.
\newblock {\em Communications of the ACM}, 63(11):139--144, 2020.

\bibitem{escontrela2022adversarial}
Alejandro Escontrela, Xue~Bin Peng, Wenhao Yu, Tingnan Zhang, Atil Iscen, Ken Goldberg, and Pieter Abbeel.
\newblock Adversarial motion priors make good substitutes for complex reward functions.
\newblock In {\em 2022 IEEE/RSJ International Conference on Intelligent Robots and Systems (IROS)}, pages 25--32. IEEE, 2022.

\bibitem{DBLP:journals/corr/MaoLXLW16}
Xudong Mao, Qing Li, Haoran Xie, Raymond Y.~K. Lau, and Zhen Wang.
\newblock Multi-class generative adversarial networks with the {L2} loss function.
\newblock {\em CoRR}, abs/1611.04076, 2016.

\bibitem{article}
Xudong Mao, Qing Li, Haoran Xie, Raymond~Y.K. Lau, Zhen Wang, and Stephen~Paul Smolley.
\newblock On the effectiveness of least squares generative adversarial networks.
\newblock {\em IEEE Transactions on Pattern Analysis \& Machine Intelligence}, 41(12):2947--2960, 2019.

\bibitem{bazarevsky2020blazepose}
Valentin Bazarevsky, Ivan Grishchenko, Karthik Raveendran, Tyler Zhu, Fan Zhang, and Matthias Grundmann.
\newblock Blazepose: On-device real-time body pose tracking.
\newblock {\em arXiv preprint arXiv:2006.10204}, 2020.

\bibitem{6512060}
Guido Gioioso, Gionata Salvietti, Monica Malvezzi, and Domenico Prattichizzo.
\newblock Mapping synergies from human to robotic hands with dissimilar kinematics: An approach in the object domain.
\newblock {\em IEEE Transactions on Robotics}, 29(4):825--837, 2013.

\bibitem{GENG2011272}
Tao Geng, Mark Lee, and Martin Hülse.
\newblock Transferring human grasping synergies to a robot.
\newblock {\em Mechatronics}, 21(1):272--284, 2011.

\bibitem{7139991}
Raúl Suárez, Jan Rosell, and Néstor García.
\newblock Using synergies in dual-arm manipulation tasks.
\newblock In {\em 2015 IEEE International Conference on Robotics and Automation (ICRA)}, pages 5655--5661, 2015.

\bibitem{isaac}
NVIDIA’s physics simulation environment for reinforcement~learning research.
\newblock Isaac gym - preview release.
\newblock \url{https://developer.nvidia.com/isaac-gym}, 2023.

\bibitem{aronov2006frechet}
Boris Aronov, Sariel Har-Peled, Christian Knauer, Yusu Wang, and Carola Wenk.
\newblock Fr{\'e}chet distance for curves, revisited.
\newblock In {\em Algorithms--ESA 2006: 14th Annual European Symposium, Zurich, Switzerland, September 11-13, 2006. Proceedings 14}, pages 52--63. Springer, 2006.

\end{thebibliography}

\end{document}